\documentclass[10pt,twocolumn,letterpaper]{article}

\usepackage{wacv}              

\definecolor{wacvblue}{rgb}{0.21,0.49,0.74}
\usepackage[pagebackref,breaklinks,colorlinks,allcolors=wacvblue]{hyperref}

\def\wacvPaperID{1782} 
\def\confName{WACV}
\def\confYear{2026}

\usepackage{algorithm}
\usepackage{algpseudocode}
\usepackage{amsmath}
\usepackage{amssymb}

\usepackage{booktabs}
\usepackage{multirow}
\usepackage{comment}

\title{Selective, Controlled and Domain-Agnostic Unlearning in Pretrained CLIP: A Training- and Data-Free Approach}

\author{Ashish Mishra, Gyanaranjan Nayak, Tarun Kumar, Arpit Shah, Suparna Bhattacharya, Martin Foltin\\
Hewlett Packard Labs\\
{\tt\small \{ashish.mishra,tarun.kumar2,arpit.shah,suparna.bhattacharya,martin.foltin\}@hpe.com}
}

\begin{document}
\maketitle
\begin{abstract}
Pretrained models like CLIP have demonstrated impressive zero-shot classification capabilities across diverse visual domains, spanning natural images, artistic renderings, and abstract representations. However, real-world applications often demand the removal (or ``unlearning'') of specific object classes without requiring additional data or retraining, or affecting the model’s performance on unrelated tasks. In this paper, we propose a novel training- and data-free unlearning framework that enables three distinct forgetting paradigms: (1) global unlearning of selected objects across all domains, (2) domain-specific knowledge removal (e.g., eliminating sketch representations while preserving photo recognition), and (3) complete unlearning in selective domains. By leveraging a multimodal nullspace through synergistic integration of text prompts and synthesized visual prototypes derived from CLIP's joint embedding space, our method efficiently removes undesired class information while preserving the remaining knowledge. This approach overcomes the limitations of existing retraining-based methods and offers a flexible and computationally efficient solution for controlled model forgetting.
\end{abstract}
    
\section{Introduction}
\label{sec:intro}
Recent advancements in multi-modal learning have led to the development of models like CLIP \cite{radford2021learning}, which exhibit remarkable zero-shot classification capabilities across diverse domains, including natural images and artistic renderings \cite{novack2023chils, qian2023intra, qian2024online}. These models, trained on extensive web-scale data, have unlocked unprecedented generalization abilities \cite{cheng2024transfer, vidit2023clip}. However, practical deployment scenarios often require selectively unlearning specific objects in certain domains while preserving them in others. For instance, ethical considerations \cite{liu2024machine, hine2024supporting}, regulatory compliance \cite{cooper2024machine, viswanath2024machine}, and dataset biases \cite{zhou2024limitations} may require the removal of certain classes from a model's predictions.

A particularly challenging aspect of unlearning arises in domain-specific contexts: a model may need to forget an object class in one domain (e.g., cartoons or sketches) while retaining it in another (e.g., real-world photographs). Traditional model editing approaches, such as fine-tuning or retraining with adjusted datasets, are often impractical due to high computational costs, reliance on extensive forget/retain data, and potential interference with unrelated model capabilities \cite{machine-unlearning-survey}. Moreover, these methods risk catastrophic forgetting of essential knowledge, leading to unintended performance degradation.

The growing prevalence of pre-trained CLIP models, trained on diverse and expansive datasets, positions them as universal models capable of addressing tasks across multiple domains. However, while object unlearning has been well-explored in single-domain settings, its application in multi-domain environments remains largely uncharted, presenting a unique challenge: handling both the object and domain dimensions simultaneously. This challenge manifests in scenarios where an object must be unlearned in specific domains while being retained in others, universally unlearned across all domains, or entirely forgotten in a particular domain. A zero-shot, data-free, and fine-tuning free approach to selective object unlearning is not only technically demanding but also critical for continual domain adaptation models. Such an approach would enhance the flexibility and efficiency of deploying CLIP models in real-world scenarios, where dynamic adaptation without the overhead of retraining is increasingly vital.

In this work, we propose a novel, data-free, and training-free approach for class-specific and domain-specific unlearning in CLIP. Rather than relying on retraining, we leverage the model's inherent multi-modal nature to construct an augmented \textit{forget subspace}. Our method synthesizes visual prototypes through gradient-based optimization and combines them with text embeddings derived from class names. This enriched representation captures the semantic and visual essence of the target class and selective domains, which we then eliminate using a closed-form nullspace projection. Our approach is efficient, interpretable, and minimally intrusive, ensuring that the model's remaining capabilities remain intact.
\begin{figure}
    \centering
    \includegraphics[width=\linewidth]{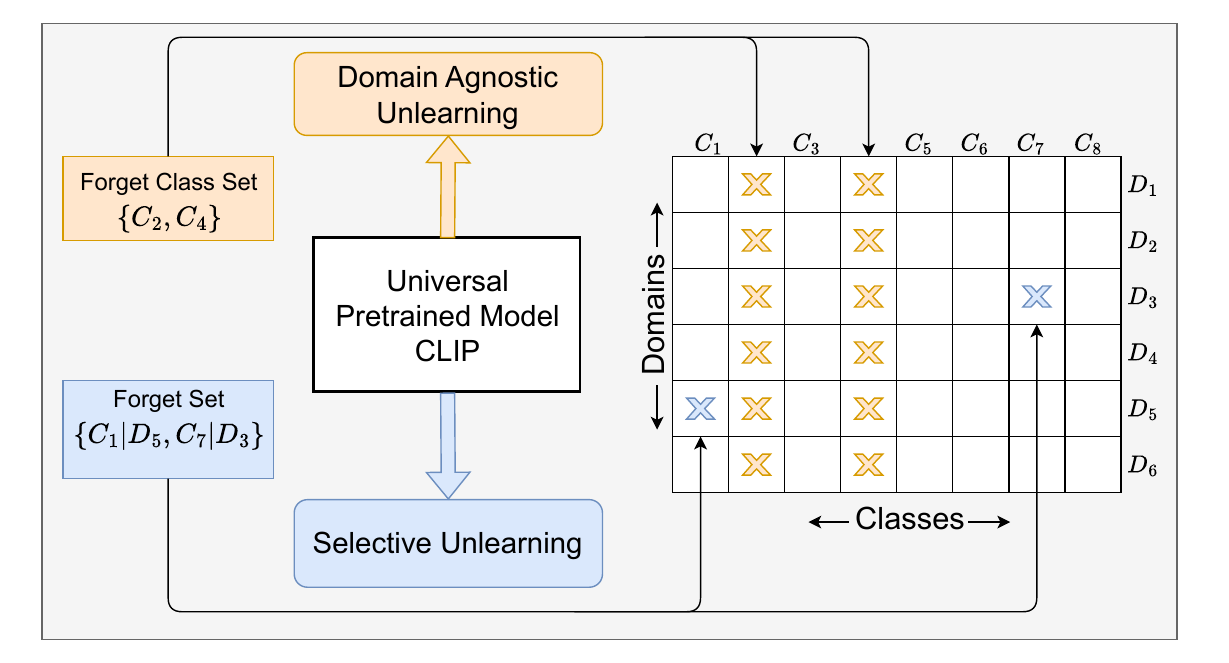}
    \caption{\textbf{Domain agnostic and Selective unlearning on CLIP model:} As shown in the (domain, class) matrix, our proposed approach can perform selective as well as domain agnostic unlearning without any fine-tuning/re-training stage.}
    \label{fig:intro}
\end{figure}

Figure \ref{fig:intro} illustrates the functioning of our proposed method to unlearn class information in a domain-agnostic or selective way. Our contributions are threefold:
\begin{enumerate}
    \item Data- and training-free unlearning: We introduce a novel method that removes object classes from CLIP without requiring additional data or retraining.
    \item Flexible, domain-specific unlearning: Our approach enables both global unlearning (across all domains) and selective unlearning in specific domains, providing greater control over model behavior.
    \item Interpretable closed-form solution: We formulate unlearning as a structured nullspace projection using singular value decomposition (SVD), ensuring precise removal of the target class from selective domains with minimal disruption to the model's overall functionality.
\end{enumerate}
Our method offers a practical and theoretically grounded solution to the unlearning problem, presenting an efficient alternative to conventional model editing techniques. Extensive experiments demonstrate its effectiveness in removing unwanted concepts while preserving the model's generalization abilities.

\section{Related Work}
Early approaches to machine unlearning in computer vision focused on exact unlearning through dataset modification and retraining, as exemplified by the SISA framework \cite{bourtoule2021machine}  that partitions data into shards and selectively retrains affected components. While effective for small-scale models, these methods face prohibitive computational costs when applied to modern architectures like Vision Transformers \cite{zhou2024limitations}. Approximate unlearning techniques emerged to address scalability issues, such as gradient ascent on targeted class embeddings or influence function-based parameter updates \cite{xu2024machine}. However, these methods depend on access to original training data for calculating Hessian matrices or identifying influential samples—a requirement incompatible with real-world scenarios where data retention is constrained by privacy laws. Domain-specific unlearning frameworks like episodic feature disentanglement attempted to remove biases in image classifiers but introduced reconstruction networks that increased parameter counts \cite{xu2024machineunlearning}. Crucially, these vision-centric approaches operate exclusively in pixel space, making them ineffective for multimodal architectures like CLIP that require joint visual-textual knowledge removal \cite{chakraborty2024can}.

Recent work has extended unlearning concepts to multimodal systems through diffusion model interventions. Erased Stable Diffusion \cite{gandikota2023erasing} removes unwanted concepts from diffusion models by a process involving fine-tuning. Concept unlearning techniques such as EraseDiff and SalUn demonstrate partial success in removing artistic styles or objects from text-to-image models but depend on anchor prompts (e.g., "vanilla photo") to guide the forgetting process—a strategy that fails when target concepts lack natural semantic counterparts \cite{zhu2024choose, zhang2024generate}. Despite claims of efficiency, diffusion unlearning methods still require a significant compute time, making them impractical for several real-world use cases.

The unique architecture of CLIP—which jointly embeds images and text into a shared latent space—has prompted specialized unlearning approaches. Prompt engineering techniques inspired by ODG-CLIP show preliminary success in domain-specific forgetting through style-aware text modifiers, yet struggle with cross-domain generalization \cite{chakraborty2024can}. A recent work CLIPErase \cite{yang2024ULcliperase} proposes a modular approach to disentangle the associations between text and image features of the \textit{forget-set}. However, its reliance on \textit{retain set} data for the Retention Module limits its applicability in several real-world scenarios. \cite{kravets2025ULzeroshotclip0} introduces a projection matrix adjustment method that operates without visual data. However, this approach risks the cross-modal alignment of retained classes due to its exclusive focus on textual embeddings. We use this method as a baseline in our experiments. We use another baseline \cite{chen2024ULUNSC} that is based on machine unlearning via Null Space Calibration (UNSC). As shown in the later sections, our proposed method outperforms the baselines and overcomes the limitations of the other approaches.  

\section{Methodology}

In this section, we present our approach for domain unlearning from a pretrained CLIP model. Our goal is to selectively remove the representation of target (forget) object classes from specified domains while preserving the class information in remaining domains. We describe the problem definition, the mathematical formulation for computing an augmented nullspace projection in three distinct setups, and detail the training and inference procedures in this section.

\subsection{Problem Definition}

Given a pretrained CLIP model, our objective is to selectively unlearn a set of target object classes (\textit{forget} classes) in designated unlearning domains, while maintaining the integrity of the \textit{retain} classes. We consider three setups:

\begin{enumerate}[label=(\arabic*), noitemsep]
    \item \textbf{Global Unlearning:} Remove the target classes uniformly across all domains (e.g., remove the Dog class from all domains such as sketch, art, cartoon, etc.). The augmented nullspace projection is computed using the embeddings of the forget classes aggregated over all available domains.
    \item \textbf{Selective Domain Unlearning:} Unlearn the target classes only in specified unlearning domains (e.g., sketches or cartoons), while preserving their representation in other domains. The augmented matrix is computed using domain-specific embeddings.
    \item \textbf{Complete Selective Domain Unlearning:} Fully eliminate any residual domain-specific signals of the target classes in the unlearning domains. This setup further refines the augmented matrix by incorporating additional domain-residual information.
\end{enumerate}

In all setups, our method targets only the forget classes, leaving the retain class representations unaffected even within unlearning domains.

\subsection{Mathematical Formulation for Domain Unlearning}

Let the CLIP model map an image \( x \) to an embedding via
\[
h = f(x;\theta) \, W,
\]
where \( f(x;\theta) \in \mathbb{R}^{1 \times D} \) denotes the pre-projection features from the visual encoder, and \( W \in \mathbb{R}^{D \times 512} \) is the final projection matrix. The text encoder maps a prompt \( p \) to a normalized text embedding:
\[
t = \frac{\text{Enc}_\text{text}(p)}{\|\text{Enc}_\text{text}(p)\|} \in \mathbb{R}^{512}.
\]

\begin{figure}
    \centering
    \includegraphics[width=\linewidth]{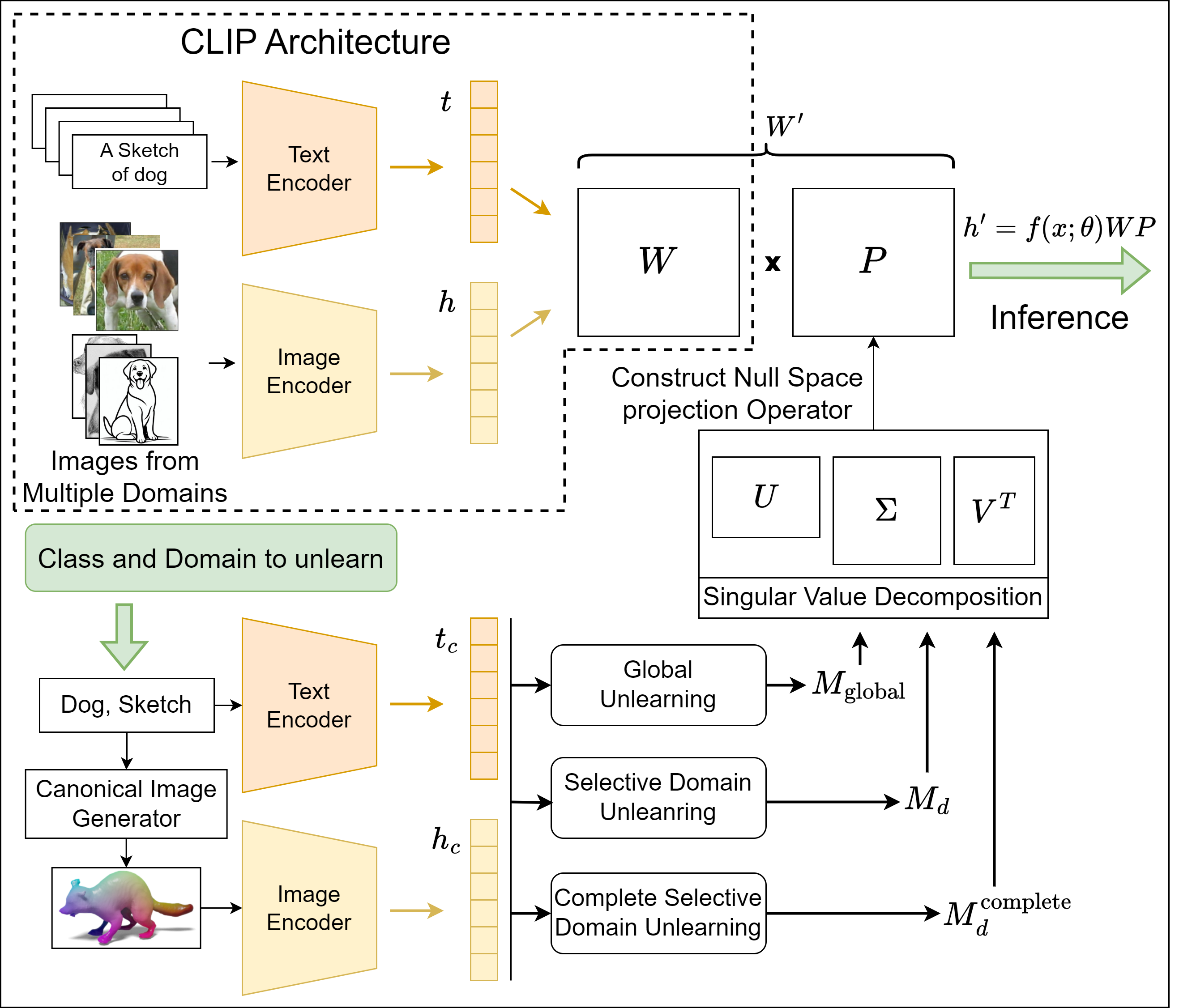}
    \caption{\textbf{Domain agnostic and Selective unlearning on CLIP model:} As shown in the (domain, class) matrix, our proposed approach can perform selective as well as domain agnostic unlearning without any fine-tuning/re-training stage.}
    \label{fig:method}
\end{figure}

As shown in Figure \ref{fig:method} For a given target (forget) object class \( c \), the following steps are performed:

\begin{enumerate}[noitemsep]
    \item \textbf{Text Embedding:} Generate the text embedding \( t_c \) using a prompt based solely on the class name (e.g., “a photo of a $c$”).
    \item \textbf{Canonical Visual Embedding:}
    \begin{itemize}[noitemsep]
        \item \textbf{Canonical Image Creation:} Synthesize a \emph{canonical image} \( x_c \) via gradient-based optimization that maximizes the cosine similarity between its visual embedding and the text embedding \( t_c \):
        \[
        \cos\left(f(x_c;\theta) \, W, \, t_c\right) = \frac{(f(x_c;\theta) \, W) \cdot t_c}{\|f(x_c;\theta) \, W\|\|t_c\|}.
        \]
        This process yields a representative image capturing the salient features of the target class.
        \item \textbf{Visual Embedding:} Compute the visual embedding of the synthesized image:
        \[
        h_c = f(x_c;\theta) \, W.
        \]
    \end{itemize}
\end{enumerate}

For a set of \( k \) target classes \( \{ c_1, c_2, \dots, c_k \} \), the computation of the augmented matrix differs according to the following setups:

\subsubsection*{(1) Global Unlearning}

In the global unlearning scenario, we aggregate the information from all domains. The augmented matrix is defined as:
\[
M_{\text{global}} = \begin{bmatrix} t_{c_1} \\ \vdots \\ t_{c_k} \\ h_{c_1} \\ \vdots \\ h_{c_k} \end{bmatrix} \in \mathbb{R}^{2k \times 512}.
\]
Taking its transpose,
\[
\tilde{M}_{\text{global}} = M_{\text{global}}^T \in \mathbb{R}^{512 \times 2k},
\]
we perform singular value decomposition (SVD):
\[
\tilde{M}_{\text{global}} = U_{\text{global}} \Sigma_{\text{global}} V_{\text{global}}^\top.
\]
An orthonormal basis \( U_{\text{global}} \) is then used to construct the nullspace projection operator:
\[
P_{\text{global}} = I_{512} - U_{\text{global}} U_{\text{global}}^\top.
\]

\subsubsection*{(2) Selective Domain Unlearning}

For selective domain unlearning, we compute domain-specific embeddings. Let \( d \) denote the designated unlearning domain. In this case, the canonical images are synthesized (or sampled) specifically for domain \( d \), yielding domain-specific visual embeddings \( h_c^d \) for each target class. The augmented matrix for domain \( d \) is:
\[
M_d = \begin{bmatrix} t_{c_1} \\ \vdots \\ t_{c_k} \\ h_{c_1}^d \\ \vdots \\ h_{c_k}^d \end{bmatrix} \in \mathbb{R}^{2k \times 512},
\]
with its transpose given by:
\[
\tilde{M}_d = M_d^T \in \mathbb{R}^{512 \times 2k}.
\]
Performing SVD:
\[
\tilde{M}_d = U_d \Sigma_d V_d^\top,
\]
we extract the orthonormal basis \( U_d \) and define the nullspace projection operator:
\[
P_d = I_{512} - U_d U_d^\top.
\]

\subsubsection*{(3) Complete Selective Domain Unlearning}

In the complete selective domain unlearning setup, we further refine the projection by incorporating any residual domain-specific information that may be present in the forget classes. In addition to the text embeddings \( t_c \) and domain-specific canonical embeddings \( h_c^d \), we introduce a residual embedding \( r_c^d \) capturing domain-specific nuances. The augmented matrix is then formed as:
\[
M_d^{\text{complete}} = \begin{bmatrix} t_{c_1} \\ \vdots \\ t_{c_k} \\ h_{c_1}^d \\ \vdots \\ h_{c_k}^d \\ r_{c_1}^d \\ \vdots \\ r_{c_k}^d \end{bmatrix} \in \mathbb{R}^{3k \times 512},
\]
and its transpose is:
\[
\tilde{M}_d^{\text{complete}} = \left(M_d^{\text{complete}}\right)^T \in \mathbb{R}^{512 \times 3k}.
\]
After performing SVD:
\[
\tilde{M}_d^{\text{complete}} = U_d^{\text{complete}} \Sigma_d^{\text{complete}} \left(V_d^{\text{complete}}\right)^T,
\]
we extract the basis \( U_d^{\text{complete}} \) and form the nullspace projection operator:
\[
P_d^{\text{complete}} = I_{512} - U_d^{\text{complete}} \left(U_d^{\text{complete}}\right)^T.
\]
This refined operator is designed to fully remove both the standard and residual domain-specific signals associated with the forget classes.

\subsection{Training (Unlearning) Process}

Our unlearning procedure is performed in a closed-form manner without modifying the original model parameters through gradient updates. The steps are as follows:

\begin{enumerate}[noitemsep]
    \item \textbf{Feature Extraction:} For each target class \( c \) (i.e., the forget classes), compute the normalized text embedding \( t_c \) and synthesize the canonical image \( x_c \) to obtain the visual embedding \( h_c \). For domain-specific setups, obtain \( h_c^d \) (and \( r_c^d \) for complete unlearning) as needed.
    \item \textbf{Subspace Construction:} Depending on the setup, construct the corresponding augmented matrix:
    \begin{itemize}[noitemsep]
        \item Global: \( M_{\text{global}} \)
        \item Selective Domain: \( M_d \)
        \item Complete Selective Domain: \( M_d^{\text{complete}} \)
    \end{itemize}
    \item \textbf{SVD and Nullspace Projection:} Compute the transpose of the augmented matrix and perform SVD to extract the orthonormal basis, and then construct the appropriate nullspace projection operator:
    \[
    P = \begin{cases}
        P_{\text{global}} & \text{(Global Unlearning)} \\
        P_d & \text{(Selective Domain Unlearning)} \\
        P_d^{\text{complete}} & \text{(Complete Selective Domain Unlearning)}
    \end{cases}
    \]
    \item \textbf{Projection Update:} Update the final projection matrix for the designated domain (or globally) as:
    \[
    W' = W \, P.
    \]
    This update effectively removes the contribution of the forget classes from the CLIP model in the specified unlearning setting while preserving the retain classes.
\end{enumerate}

\subsection{Inference Process}

At inference time, the embedding computation depends on the domain of the input image:

\begin{itemize}[noitemsep]
    \item \textbf{For Images from Unlearning Domains:} If an input image \( x \) originates from a domain where unlearning has been applied, its embedding is computed using the updated projection matrix:
    \[
    h' = f(x;\theta) \, W' = f(x;\theta) \, W \, P.
    \]
    This adjustment reduces the cosine similarity between \( h' \) and the target class embeddings.
    \item \textbf{For Images from Other Domains:} If the image comes from a domain not targeted for unlearning, the original projection matrix \( W \) is used:
    \[
    h = f(x;\theta) \, W.
    \]
\end{itemize}

\section{Experimental Setup and Results}

\begin{table*}[ht]
\centering
\caption{CLIP Model Accuracy on DomainNet Dataset: Pre-Unlearning vs. Post-Unlearning across Selective Forget Domain Combinations (6 Evaluation Domains) with Additional Lip Baseline}
\resizebox{\textwidth}{!}{%
\begin{tabular}{ll|cccccccc|cccccccc}
\toprule
\textbf{Selective Forget} & \textbf{Domain}
  & \multicolumn{8}{c|}{\textbf{Retain Set Accuracy (\%)}} 
  & \multicolumn{8}{c}{\textbf{Forget Set Accuracy (\%)}} \\
\cmidrule(lr){3-10} \cmidrule(lr){11-18}
& & \multicolumn{2}{c}{\textbf{NSC}} & \multicolumn{2}{c}{\textbf{ZSL-CLIP}} & \multicolumn{2}{c}{\textbf{Lip}} & \multicolumn{2}{c|}{\textbf{Ours}}
  & \multicolumn{2}{c}{\textbf{NSC}} & \multicolumn{2}{c}{\textbf{ZSL-CLIP}} & \multicolumn{2}{c}{\textbf{Lip}} & \multicolumn{2}{c}{\textbf{Ours}} \\
\cmidrule(lr){3-4} \cmidrule(lr){5-6} \cmidrule(lr){7-8} \cmidrule(lr){9-10}
\cmidrule(lr){11-12} \cmidrule(lr){13-14} \cmidrule(lr){15-16} \cmidrule(lr){17-18}
& & BF & AF & BF & AF & BF & AF & BF & AF
  & BF & AF & BF & AF & BF & AF & BF & AF \\
\midrule
\multirow{6}{*}{clipart}
   & clipart    & 91.10 & 87.50 & 91.10 & 90.51 & 90.50 & 89.90 & 91.10 & 87.03 & 95.98 & 20.30 & 95.98 & 12.20 & 95.30 & 12.00 & 95.98 & 4.51 \\
   & infograph  & 69.10 & 67.50 & 69.10 & 69.00 & 68.60 & 68.20 & 69.10 & 69.10 & 71.95 & 16.06 & 71.95 & 10.50 & 71.60 & 10.30 & 71.95 & 71.95 \\
   & painting   & 80.80 & 78.51 & 80.80 & 80.85 & 80.30 & 80.15 & 80.80 & 80.70 & 86.87 & 18.30 & 86.87 & 8.41 & 86.30 & 8.11 & 86.87 & 86.50 \\
   & quickdraw  & 38.45 & 34.13 & 38.45 & 36.60 & 38.10 & 36.00 & 38.45 & 38.40 & 46.03 & 18.90 & 46.03 & 11.30 & 45.80 & 11.00 & 46.03 & 46.00 \\
   & real       & 95.97 & 95.02 & 95.97 & 95.80 & 95.70 & 95.40 & 95.97 & 95.97 & 96.36 & 27.10 & 96.36 & 7.05 & 96.05 & 6.80 & 96.36 & 96.36 \\
   & sketch     & 89.04 & 64.46 & 89.04 & 86.20 & 88.50 & 85.90 & 89.04 & 89.00 & 87.76 & 11.14 & 87.76 & 15.80 & 87.30 & 15.20 & 87.76 & 87.70 \\
\midrule
\multirow{6}{*}{painting}
   & clipart    & 91.10 & 87.50 & 91.10 & 90.51 & 90.60 & 90.00 & 91.10 & 91.10 & 95.98 & 20.30 & 95.98 & 12.20 & 95.40 & 11.80 & 91.76 & 91.76 \\
   & infograph  & 69.10 & 67.50 & 69.10 & 69.00 & 68.80 & 68.00 & 69.10 & 69.10 & 71.95 & 16.06 & 71.95 & 10.50 & 71.40 & 10.00 & 71.95 & 71.95 \\
   & painting   & 80.80 & 78.51 & 80.80 & 80.85 & 80.25 & 79.80 & 80.80 & 78.57 & 86.87 & 18.30 & 86.87 & 8.41 & 86.10 & 8.01 & 86.87 & 3.51 \\
   & quickdraw  & 38.45 & 34.13 & 38.45 & 36.60 & 38.10 & 35.90 & 38.45 & 38.45 & 46.03 & 18.90 & 46.03 & 11.30 & 45.65 & 10.90 & 46.00 & 46.03 \\
   & real       & 95.97 & 95.02 & 95.97 & 95.80 & 95.70 & 95.30 & 95.97 & 95.97 & 96.36 & 27.10 & 96.36 & 7.05 & 96.05 & 6.60 & 96.36 & 96.36 \\
   & sketch     & 89.04 & 64.46 & 89.04 & 86.20 & 88.80 & 85.20 & 89.02 & 89.04 & 87.76 & 11.14 & 87.76 & 15.80 & 87.15 & 15.30 & 85.71 & 85.76 \\
\midrule
\multirow{6}{*}{painting+real+sketch}
   & clipart    & 91.10 & 87.50 & 91.10 & 90.51 & 90.80 & 90.10 & 91.10 & 91.10 & 95.98 & 20.30 & 95.98 & 12.20 & 95.30 & 12.00 & 91.76 & 91.76 \\
   & infograph  & 69.10 & 67.50 & 69.10 & 69.00 & 68.70 & 68.20 & 69.10 & 69.10 & 71.95 & 16.06 & 71.95 & 10.50 & 71.50 & 10.00 & 71.95 & 71.95 \\
   & painting   & 80.80 & 78.51 & 80.80 & 80.85 & 80.20 & 79.00 & 80.80 & 76.87 & 86.87 & 18.30 & 86.87 & 8.41 & 86.15 & 8.05 & 86.87 & 3.07 \\
   & quickdraw  & 38.45 & 34.13 & 38.45 & 36.60 & 38.00 & 35.90 & 38.45 & 38.45 & 46.03 & 18.90 & 46.03 & 11.30 & 45.70 & 10.90 & 46.03 & 46.03 \\
   & real       & 95.97 & 95.02 & 95.97 & 95.80 & 95.70 & 94.90 & 95.97 & 94.14 & 96.36 & 27.10 & 96.36 & 7.05 & 96.06 & 6.75 & 96.36 & 2.27 \\
   & sketch     & 89.04 & 64.46 & 89.04 & 86.20 & 88.70 & 85.20 & 89.04 & 85.30 & 87.76 & 11.14 & 87.76 & 15.80 & 87.20 & 15.00 & 85.76 & 2.64 \\
\midrule
\multirow{6}{*}{clipart+sketch}
   & clipart    & 91.10 & 87.50 & 91.10 & 90.51 & 90.60 & 89.50 & 91.10 & 88.16 & 95.98 & 20.30 & 95.98 & 12.20 & 95.20 & 12.00 & 91.76 & 3.11 \\
   & infograph  & 69.10 & 67.50 & 69.10 & 69.00 & 68.80 & 68.00 & 69.10 & 69.10 & 71.95 & 16.06 & 71.95 & 10.50 & 71.40 & 10.10 & 71.95 & 71.95 \\
   & painting   & 80.80 & 78.51 & 80.80 & 80.85 & 80.20 & 80.00 & 80.80 & 80.80 & 86.87 & 18.30 & 86.87 & 8.41 & 86.25 & 8.10 & 86.87 & 86.87 \\
   & quickdraw  & 38.45 & 34.13 & 38.45 & 36.60 & 38.00 & 35.80 & 38.45 & 38.45 & 46.03 & 18.90 & 46.03 & 11.30 & 45.75 & 11.10 & 46.03 & 46.03 \\
   & real       & 95.97 & 95.02 & 95.97 & 95.80 & 95.80 & 95.00 & 95.97 & 95.97 & 96.36 & 27.10 & 96.36 & 7.05 & 96.20 & 6.95 & 96.36 & 96.36 \\
   & sketch     & 89.04 & 64.46 & 89.04 & 86.20 & 88.80 & 85.00 & 89.04 & 85.19 & 87.76 & 11.14 & 87.76 & 15.80 & 87.20 & 15.10 & 85.76 & 2.36 \\
\bottomrule

\end{tabular}
}
\label{tab:tab1}
\end{table*}

\begin{table*}[ht]
\centering
\caption{CLIP Model Accuracy on PACS Dataset: Pre-Unlearning vs. Post-Unlearning across Selective Forget Domain Combinations (with MIA Score)}
\begin{tabular}{llcccc|c}
\toprule
\textbf{Selective Forget} & \textbf{Domain} 
& \multicolumn{2}{c}{\textbf{Retain Set Accuracy (\%)}} 
& \multicolumn{2}{c}{\textbf{Forget Set Accuracy (\%)}} 
& \textbf{MIA} \\
\cmidrule(lr){3-4} \cmidrule(lr){5-6}
\textbf{Domains} & & Pre-Unlearning & Post-Unlearning & Pre-Unlearning & Post-Unlearning & \\
\midrule
\multirow{4}{*}{Photo} 
   & Art Painting & 95.98 & 95.98 & 97.64 & 97.64 & 0.00 \\
   & Cartoon      & 97.64 & 97.64 & 99.70 & 99.70 & 0.00 \\
   & Photo        & 99.89 & 99.78 & 100.00 & 5.01 & \textbf{94.88} \\
   & Sketch       & 99.34 & 99.34 & 93.04 & 93.04 & 0.00 \\
\midrule
\multirow{4}{*}{Sketch} 
   & Art Painting & 95.98 & 95.98 & 97.64 & 97.64 & 0.00 \\
   & Cartoon      & 97.64 & 97.64 & 99.70 & 99.70 & 0.00 \\
   & Photo        & 99.89 & 99.89 & 100.00 & 100.00 & 0.00 \\
   & Sketch       & 99.34 & 99.34 & 93.34 & 12.95 & \textbf{86.05} \\
\midrule
\multirow{4}{*}{Photo + Sketch} 
   & Art Painting & 95.98 & 95.98 & 97.60 & 97.64 & 0.04 \\
   & Cartoon      & 97.64 & 97.60 & 99.70 & 99.70 & 0.04 \\
   & Photo        & 99.89 & 99.78 & 99.91 & 4.48 & \textbf{95.08} \\
   & Sketch       & 99.34 & 99.24 & 93.30 & 15.68 & \textbf{78.46} \\
\midrule
\multirow{4}{*}{Photo + Sketch + Cartoon} 
   & Art Painting & 98.90 & 95.98 & 97.64 & 97.64 & -2.92 \\
   & Cartoon      & 97.64 & 97.05 & 99.70 & 15.60 & \textbf{84.99} \\
   & Photo        & 99.89 & 99.78 & 100.00 & 5.80 & \textbf{95.31} \\
   & Sketch       & 99.34 & 99.43 & 93.34 & 7.55 & \textbf{85.68} \\
\bottomrule
\end{tabular}
\label{tab:tab2}
\end{table*}

\begin{table*}[ht]
\centering
\caption{CLIP Model Accuracy on PACS Dataset: Domain Agnostic Unlearning (Forget Set Classes Unlearned from All Domains)}
\begin{tabular}{lcccc}
\toprule
\textbf{Domain} & \multicolumn{2}{c}{\textbf{Retain Set Accuracy (\%)}} & \multicolumn{2}{c}{\textbf{Forget Set Accuracy (\%)}} \\
\cmidrule(lr){2-3} \cmidrule(lr){4-5}
                & Pre-Unlearning & Post-Unlearning & Pre-Unlearning & Post-Unlearning \\
\midrule
Art Painting    & 95.98             & 95.77             & 97.64             & 20.11             \\
Cartoon         & 97.64             & 96.95             & 99.70             & 18.40             \\
Photo           & 99.89             & 99.78             & 100             & 22.53             \\
Sketch          & 99.34             & 99.72             & 93.04             & 17.63             \\
\bottomrule
\end{tabular}
\label{tab:tab3}
\end{table*}

\subsection{Datasets} We evaluated our proposed approach using two widely adopted datasets for domain adaptation: PACS \cite{PACS} and DomainNet \cite{peng2019moment}. The PACS dataset comprises four domains—Art Painting, Cartoon, Photo, and Sketch—with each domain containing images from seven object classes. DomainNet is a more challenging dataset that includes six different domains—Clipart, Infograph, Painting, Quickdraw, Real, and Sketch—with each domain containing images for X classes.

\subsection{Evaluation} We assessed our model using three different setups:
\begin{itemize}
    \item \textbf{Domain Agnostic Unlearning:} In this configuration, the classes to be forgotten are removed from all domains, while the classes that are retained remain preserved in every domain. By simply providing the names of the classes to forget, our approach eliminates their information across all domains.
    \item \textbf{Selective Domain Unlearning:} In this setup, we specify a particular domain from which the classes should be unlearned by the CLIP model. Meanwhile, the other domains continue to retain the information for these classes.
    \item \textbf{Complete Selective Domain Unlearning:} Here, we only indicate the domain from which to unlearn, and as a result, all information related to that domain is entirely erased from the CLIP model. The same class information, however, remains intact in the other domains.
\end{itemize}

\subsection{Datasets Split}
For the PACS dataset, which comprises seven classes, we divided them into a forget set with three classes and a retain set with four classes, and evaluated our approach using selective domain unlearning in both single-domain and multi-domain setups.

For the domainNet dataset, we randomly selected 40 identical classes from each domain to construct our dataset, and then randomly split these into a target set and a retain set, each containing 20 classes, to evaluate our approach.

\subsection{Evaluation Metrics}
We evaluated our approach by measuring class accuracy for both the target/forget set classes and the retain set classes, comparing performance before forgetting/unlearning (BF) and after forgetting/unlearning (AF) across the selective unlearning domains. To evaluate selective domain unlearning, we additionally report the MIA score. A high MIA score demonstrates that the method can effectively remove knowledge of the forget set classes from the unlearned domain, while maintaining high accuracy on the retain set classes from the remaining domains.

We evaluated the effectiveness of our approach using the Membership Inference Attack (MIA) score. In the context of machine unlearning, the MIA score quantifies how well a model "forgets" information about data designated for removal (the "forget set") while continuing to retain knowledge about the remaining data (the "retain set"). We calculated the MIA as:
\begin{equation}
\text{MIA} = \left( \text{BF}_{\text{forget}} - \text{AF}_{\text{forget}} \right)
           - \left( \text{BF}_{\text{retain}} - \text{AF}_{\text{retain}} \right)
\end{equation}
A high MIA score indicates a more effective unlearning approach, as it shows that the model has successfully forgotten the designated classes while retaining knowledge of the remaining classes.

\subsection{Baseline}
Unlearning of target objects across multiple selective domains in pretrained CLIP models remains underexplored. Although ZSL-CLIP \cite{kravets2025ULzeroshotclip0} and Lip \cite{lip} have addressed target class unlearning within single domains, it does not extend to multiple domains. Consequently, we adopted the ZSL-CLIP and Lip approaches as baselines to assess our method's efficacy and effectiveness. Additionally, we included Machine Unlearning via Null Space Calibration (NSP) \cite{chen2024ULUNSC}, originally designed for class unlearning, as another baseline because it aligns well with our work.

To demonstrate the generalizability of our approach, we evaluated it on two versions of CLIP: CLIP-ViT/16 and CLIP-ViT/32. The main manuscript includes all results for CLIP-ViT/32, while the results for CLIP-ViT/16 are provided in the supplementary material.

\section{Results}
To demonstrate the effectiveness of our approach, we conducted experiments on two widely used domain adaptation datasets—PACS and the more challenging DomainNet—across three distinct setups: global domain unlearning, selective domain unlearning, and complete domain unlearning. To highlight the significance of our method, we compared it with two baseline approaches: NSC and ZSL-CLIP. The detailed analysis of the results for each setup is provided below.

\subsection{Selective Domains Unlearning}
In this setup, the objective is to unlearn the designated forget class set from specific domains while retaining its information in other domains, and ensuring that the retain class set remains consistent before and after unlearning across all domains. We evaluated our approach for both single-domain unlearning and multiple selective domain unlearning. As shown in Table \ref{tab:tab1}, our method effectively unlearns the forget set in targeted domains while preserving its information in other domains, and the accuracy for the retain set remains nearly identical before and after unlearning. This demonstrates the effectiveness of our approach for selective domain unlearning in the CLIP model, and notably, our method performs well even when unlearning is applied to multiple domains simultaneously. For instance, if we aim to unlearn the photo, sketch, and cartoon domains while preserving the Art painting domain from the PACS dataset, Table \ref{tab:tab2} clearly shows that the accuracy for the forget class set in the photo, sketch, and cartoon domains drops significantly after unlearning, whereas the accuracy for the Art painting domain remains largely unchanged. Additionally, the accuracy for the retained set stays approximately the same before and after unlearning across all domains. This demonstrates the effectiveness of our approach in this CLIP model unlearning setup.

A similar trend was observed for the DomainNet dataset, as illustrated in Table \ref{tab:tab1}.

In Table \ref{tab:tab1}, we also compare our approach with two baselines, NSC and ZSL-CLIP, for selective domain unlearning on the challenging DomainNet dataset. Our method clearly outperforms both baselines in terms of forgetting/unlearning and retention accuracy. The baselines were originally designed for class unlearning regardless of domains, so they remove target classes from all domains within the CLIP model. This is because they focus solely on the object class structure and cannot distinguish between different domains, making them unsuitable for a selective domain-unlearning setup.

\subsection{Domain Agnostic Unlearning}
In this setup, only the forget class set is provided, which must be unlearned across all domains, while the retain class set is preserved. In other words, the target forget objects are entirely removed from the pretrained CLIP model's domain representations, a process we refer to as domain-agnostic unlearning. The results for this setup are presented in Table \ref{tab:tab3} for PACS and Table \ref{tab:4} for DomainNet. The data clearly shows that the forgotten classes are effectively unlearned across all domains, and the accuracy for the retained classes remains consistent before and after unlearning for both datasets. This scenario represents a special case of selective domain unlearning, where all domains are selected for unlearning.
 
\subsection{Complete Domain Unlearning}
In this setup, the goal is to completely unlearn all information related to an entire domain from the CLIP model. Instead of distinguishing between forget and retain classes, any object from the domain to be unlearned should be forgotten, while the same object remains recognizable in other domains. Table \ref{tab:tab2} presents the results for the PACS dataset, while Table \ref{tab:tab6} shows the results for DomainNet regarding real domain unlearning. Both tables report the average accuracies across all six domains before and after unlearning. The results clearly indicate that information from the unlearned domain is effectively removed, yet the same object can still be accurately recognized in the other domains.

\begin{table}[ht]
\centering
\caption{Full Domain Unlearning Evaluation: Accuracy on All Domains for Each Unlearn Domain (All Classes as Forget)}
\resizebox{0.5\textwidth}{!}{%
\begin{tabular}{l|cc|cc|cc|cc}
\toprule
\multirow{2}{*}{\textbf{Unlearn Domain}} & \multicolumn{2}{c|}{\textbf{Art Painting}} & \multicolumn{2}{c|}{\textbf{Cartoon}} & \multicolumn{2}{c|}{\textbf{Photo}} & \multicolumn{2}{c}{\textbf{Sketch}} \\
\cmidrule(lr){2-9}
 & Pre & Post & Pre & Post & Pre & Post & Pre & Post \\
\midrule
Art Painting & 99.45 & 9.65 & 97.88 & 97.48 & 100 & 99.82 & 90.70 & 85.87 \\
Cartoon      & 99.45 & 95.31 & 97.88 & 8.28 & 100 & 99.85 & 90.70 & 86.89 \\
Photo        & 99.45 & 95.31 & 97.88 & 97.48 & 100 & 2.51 & 90.70 & 88.94 \\
Sketch       & 99.45 & 99.31 & 97.88 & 97.48 & 100 & 99.82 & 90.70 & 6.16 \\
\bottomrule
\end{tabular}%
}
\label{tab:tab5}
\end{table}

\begin{table}[ht]
\centering
\caption{Full Domain Unlearning for Real Domain Evaluation: Average Accuracy on All 6 Domains}
\resizebox{\linewidth}{!}{%
\begin{tabular}{l|cccccc}
\toprule
\textbf{Unlearn Real Domain} & \textbf{clipart} & \textbf{infograph} & \textbf{painting} & \textbf{quickdraw} & \textbf{real} & \textbf{sketch} \\
\midrule
Pre-Unlearning  & 87.33 & 61.73 & 85.08 & 34.95 & 94.95 & 74.20 \\
Post-Unlearning & 87.33  & 61.73 & 85.08 & 34.95 & 0.92 & 74.20 \\
\bottomrule
\end{tabular}%
}
\label{tab:tab6}
\end{table}

\begin{table*}[ht]
\centering
\caption{CLIP Model Accuracy on DomainNet Dataset: Domain Agnostic Unlearning (Forget Set Classes Unlearned from All Domains)}
\begin{tabular}{lcccc}
\toprule
\textbf{Domain} & \multicolumn{2}{c}{\textbf{Retain Set Accuracy (\%)}} & \multicolumn{2}{c}{\textbf{Forget Set Accuracy (\%)}} \\
\cmidrule(lr){2-3} \cmidrule(lr){4-5}
                & Pre-Unlearning & Post-Unlearning & Pre-Unlearning & Post-Unlearning \\
\midrule
clipart    & 92.04          & 80.77          & 86.64          & 3.41          \\
infograph         & 77.17          & 62.5          & 70.44          & 9.95          \\
painting           & 88.83          & 75.9          & 86.76         & 2.45          \\
quickdraw          & 38.04          & 29.2          & 47.28          & 3.53          \\
real        & 97.34             & 91.82             & 93.62             & 6.09             \\
sketch        & 86.31             & 81.70             & 81.00             & 4.20             \\
\bottomrule
\end{tabular}
\label{tab:4}
\end{table*}


\begin{figure}[h]
\centering
\includegraphics[width=0.5\textwidth]{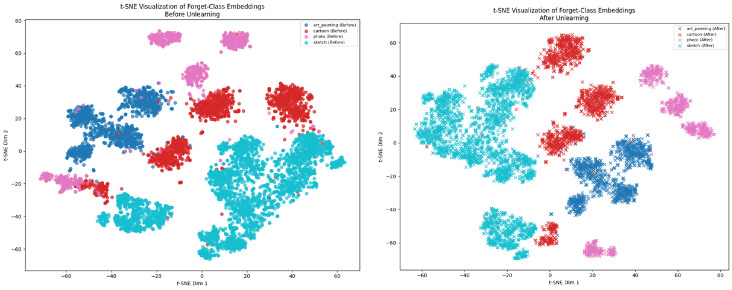}
\caption{t-SNE visualization of the projected features before and after unlearning in the photo domain.
}
\label{fig:ab1}
\end{figure}

In figure \ref{fig:ab1}, although t-SNE plots of the forget domain remain clustered after null-space projection-based unlearning, this does not indicate retained knowledge. The clustering reflects local geometric structure, while the model’s ability to utilize these features has been disrupted, demonstrating effective unlearning despite the visual appearance.

\subsection{Limitations:} A current limitation of our selective domain unlearning approach is that it requires prior knowledge of the names of the classes or domains to be forgotten. This restricts its applicability in realistic open-world scenarios, where new domains or classes may emerge over time and may not be known in advance. Addressing this limitation—by enabling the method to handle dynamically expanding or previously unseen domains—is an important direction for future work. Overcoming this challenge would significantly enhance the practical utility and robustness of our approach in real-world applications.

\section{Conclusion}
\label{sec:conclusion}
In this paper, we proposed a framework to selectively unlearn domain-specific \textit{forget class} information from the CLIP model while preserving the \textit{retain class} and information in other domains. The key contribution of our work is that this selective unlearning is achieved without requiring any retraining, fine-tuning, or access to training or forget-data. Our approach relies on a simple yet effective mechanism that computes a null space projection of the information to be forgotten and maps the final layer of CLIP to this new space.

Experimental results demonstrate that our method outperforms the baseline approach, showcasing its effectiveness in mitigating unwanted knowledge while maintaining the integrity of retained information. This work has significant implications for various AI applications where controlled forgetting is crucial, such as privacy compliance, dynamic knowledge updating, and reducing biases in pre-trained models.

Our work opens multiple avenues for further research and practical applications. One promising direction is extending this approach to other foundation models, including large language models (LLMs) and multimodal models, to enable broader applicability across different domains. Investigating the scalability of our method for handling large-scale unlearning scenarios, where multiple concepts or entire domains need to be removed without performance degradation, is another important avenue. Furthermore, integrating unlearning techniques into real-world AI pipelines could enhance compliance with evolving data privacy regulations, such as GDPR and CCPA, by enabling efficient removal of personal or sensitive data from models. Exploring efficient real-time implementations of unlearning in edge AI and federated learning settings could further expand its practical impact.

\clearpage
{
    \small
    \bibliographystyle{ieeenat_fullname}
    \bibliography{main}
}

\end{document}